\def\year{2021}\relax
\documentclass[letterpaper]{article} % DO NOT CHANGE THIS
\usepackage{aaai21}  % DO NOT CHANGE THIS
\usepackage{times}  % DO NOT CHANGE THIS
\usepackage{helvet} % DO NOT CHANGE THIS
\usepackage{courier}  % DO NOT CHANGE THIS
\usepackage[hyphens]{url}  % DO NOT CHANGE THIS
\usepackage{graphicx} % DO NOT CHANGE THIS
\usepackage{natbib}  % DO NOT CHANGE THIS AND DO NOT ADD ANY OPTIONS TO IT
\usepackage{caption} % DO NOT CHANGE THIS AND DO NOT ADD ANY OPTIONS TO IT
\usepackage{times}
\usepackage{amsmath}
\usepackage{booktabs}
\usepackage{makecell}
\usepackage{multirow}
\usepackage{pifont}

\usepackage{latexsym}
\usepackage{graphicx}
\usepackage{tikz}
\usepackage{pgfplots}
\usepackage{tikz-dependency}
\usepackage{subcaption}

\usepackage[switch]{lineno} %

\usepackage{microtype}

\usepackage{microtype}
\usepackage{bbold}
\usepackage{ulem}
\usepackage{graphicx}

\usepackage{booktabs}
\usepackage{bbm}

\usepackage{subcaption}

\title{ALP-KD: Attention-Based Layer Projection for Knowledge Distillation}
\author {
        Peyman Passban\thanks{This work has been done while Peyman Passban was at Huawei.}\hspace{3mm} %\textsuperscript{\rm 1}
        Yimeng Wu\hspace{3mm} %\textsuperscript{\rm 2}
        Mehdi Rezagholizadeh\hspace{3mm} % \textsuperscript{\rm 1}
        Qun Liu\\
}
\affiliations {
    Huawei Noah's Ark Lab\\
    
    \texttt{passban.peyman@gmail.com}\\
    \texttt{\{yimeng.wu,mehdi.rezagholizadeh,qun.liu\}@huawei.com}
}
\pgfplotsset{compat=1.16}
\begin{document}

\maketitle
%\linenumbers %
\begin{abstract}
Knowledge distillation is considered as a training and compression strategy in which two neural networks, namely a \textit{teacher} and a \textit{student}, are coupled together during training. The teacher network is supposed to be a trustworthy predictor and the student tries to mimic its predictions. Usually, a student with a lighter architecture is selected so we can achieve compression and yet deliver high-quality results. In such a setting, distillation only happens for final predictions whereas the student could also benefit from teacher's supervision for internal components. 

Motivated by this, we studied the problem of distillation for intermediate layers. Since there might not be a one-to-one alignment between student and teacher layers, existing techniques skip some teacher layers and only distill from a subset of them. This shortcoming directly impacts quality, so we instead propose a combinatorial technique which relies on attention. Our model fuses teacher-side information and takes each layer's significance into consideration, then performs distillation between combined teacher layers and those of the student. Using our technique, we distilled a $12$-layer BERT \cite{Devlin2019BERTPO} into $6$-, $4$-, and $2$-layer counterparts and evaluated them on GLUE tasks \cite{Wang2018GLUEAM}. Experimental results show that our combinatorial approach is able to outperform other existing techniques.
\end{abstract}

\section{Introduction}
Knowledge distillation (KD) \citep{bucilua2006model,hintonkd} is a commonly-used technique to reduce the size of large neural networks \citep{Sanh2019DistilBERTAD}. Apart from this, we also consider it as a complementary and generic add-on to enrich the training process of any neural model \cite{furlanello2018born}. 

In KD, a student network ($\mathcal{S}$) is glued to a powerful teacher ($\mathcal{T}$) during training. These two networks can be trained simultaneously or $\mathcal{T}$ can be a pre-trained model. Usually, $\mathcal{T}$ uses more parameters than $\mathcal{S}$ for the same task, therefore it has a higher learning capacity and is expected to provide reliable predictions. On the other side, $\mathcal{S}$ follows its teacher with a simpler architecture. For a given input, both models provide predictions where those of the student are penalized by an ordinary loss function (using \textit{hard} labels) as well as predictions received from $\mathcal{T}$ (also known as \textit{soft} labels).

Training a (student) model for a natural language processing (NLP) task can be formalized as a multi-class classification problem to minimize a cross-entropy (\textit{ce}) loss function, as shown in Equation \ref{eq:1}:
\begin{multline}
    \label{eq:1}
    \mathcal{L}_{ce} = - \sum_{i=1}^{N} \sum_{w \in V} [\mathbbm{1}(y_i=w) \times\\\log p_{_\mathcal{S}}(y_i=w | x_i, \theta_{_\mathcal{S}})]
\end{multline}
where $\mathbbm{1}(.)$ is the indicator function, $V$ is a vocabulary set (or different classes in a multi-class problem), $N$ is the number of tokens in an input sequence, and $y$ is a prediction of the network $\mathcal{S}$ with a parameter set $\theta_{_\mathcal{S}}$ given an input $x$. 

To incorporate teacher's supervision, KD accompanies $\mathcal{L}_{ce}$ with an auxiliary loss term, $\mathcal{L}_{_{KD}}$, as shown in Equation \ref{eq:2}: 
\begin{multline}\label{eq:2}
    \mathcal{L}_{_{KD}} = - \sum_{i=1}^{N} \sum_{w \in V} [p_{_\mathcal{T}}(y_i=w | x_i, \theta_{_\mathcal{T}}) \times\\\log p_{_\mathcal{S}}(y_i=w | x_i, \theta_{_\mathcal{S}})]
\end{multline}
Since $\mathcal{S}$ is trained to behave identically to $\mathcal{T}$, model compression can be achieved if it uses a simpler architecture than its teacher. However, if these two models are the same size KD would still be beneficial. What $\mathcal{L}_{_{KD}}$ proposes is an ensemble technique by which the student is informed about teacher's predictions. The teacher has better judgements and this helps the student learn how much it deviates from true labels. 

This form of KD that is referred to as Regular KD (RKD) throughout this paper, only provides $\mathcal{S}$ with external supervision for final predictions, but this can be extended to other components such as intermediate layers too. The student needs to be aware of the information flow inside teacher's layers and this becomes even more crucial when distilling from deep teachers. Different alternatives have been proposed to this end, which compare networks' internal layers in addition to final predictions \cite{jiao2019tinybert,sun2020mobilebert,pkd}, but they suffer from other types of problems. The main goal in this paper is to study such models and address their shortcomings.  

\subsection{Problem Definition}
\begin{figure}[h]
\begin{center}
\includegraphics[scale=0.5]{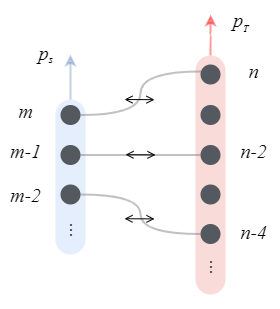}
\end{center}
\caption{\label{fig:0} Student and teacher models have $m$ and $n$ layers, respectively. Each node is an intermediate layer and links are cross-model connections. In this example, every other layer of the teacher is skipped in order to match the size of the student. The output of nodes connected to each other are compared via a loss function (shown with $\leftrightarrow$) to ensure that the student model has similar internal representations as its teacher.}
\end{figure}

To utilize intermediate layers' information (and other components in general), a family of models exists that defines a dedicated loss function to measure how much a student diverges from its teacher in terms of internal representations. In particular, if the goal is to distill from an $n$-layer teacher into an $m$-layer student, a subset of $m$ (out of $n$) teacher layers is selected whose outputs are compared to those of student layers (see Equation \ref{eq:3} for more details). Figure \ref{fig:0} illustrates this concept. 

As the figure shows, each student layer is connected to a single, dedicated peer on the teacher side, e.g. the $n$-\textit{th} teacher layer corresponds to the $m$-\textit{th} student layer. Since outputs of these two layers are compared to each other, we hope that both models generate as similar outputs as possible at points $n$ and $m$. With this simple technique, teacher's knowledge can be used to supervise student's intermediate layers. 

Experimental results show that intermediate layer matching could be quite effective, but in our study we realized that it may suffer from two shortcomings: 
\begin{itemize}
    \item If $n \gg m$, multiple layers in $\mathcal{T}$ have to be ignored for distillation but we know that those layers consist of precious information for which we spend expensive resources to learn. This issue is referred to as the \textit{skip} problem in this paper.
    
    \item Moreover, it seems the way teacher layers are kept/skipped is somewhat arbitrary as there is no particular strategy behind it. Before training, we lack enough knowledge to judge which subset of teacher layers contributes more to the distillation process, so there is a good chance of skipping significant layers if we pick them in an arbitrary fashion. Finding the best subset of layers to distill from requires an exhaustive search or an expert in the field to signify connections. We refer to this issue as the \textit{search} problem.
\end{itemize}

In order to resolve the aforementioned issues we propose an alternative, which is the main contribution of this paper. Our solution does not skip any layer but utilizes \textit{all} information stored inside $\mathcal{T}$. Furthermore, it combines teacher layers through an attention mechanism, so there is no need to deal with the search problem. We believe that the new notion of combination defined in this paper is as important as our novel KD architecture and can be adapted to other tasks too.   

The remainder of this paper is organized as follows: First, we briefly review KD techniques used in similar NLP applications, then we introduce our methodology and explain how it addresses existing shortcomings. We accompany our methodology with experimental results to show whether the proposed technique is useful. Finally, we conclude the paper and discuss future directions.

\section{Related Work}
\label{background}
KD was originally proposed for tasks other than NLP \cite{bucilua2006model,hintonkd}. \citet{kim2016sequence} adapted the idea and proposed a sequence-level extension for machine translation. \citet{freitag2017ensemble} took a step further and expanded it to a multi-task scenario. Recently, with the emergence of large  NLP and language understanding (NLU) models such as ELMO \citep{Peters:2018} and BERT \citep{Devlin2019BERTPO} KD has gained extra attention. Deep models can be trained in a better fashion and compressed via KD, which is favorable in many ways. Therefore, a large body of work in the field such as Patient KD (PKD) \citep{pkd} has been devoted to compressing/distilling BERT (and  similar) models. 

PKD is directly related to this work, so we discuss it in more detail. It proposes a mechanism to match teacher and student models' intermediate layers by defining a third loss function, $\mathcal{L}_{P}$, in addition to $ \mathcal{L}_{ce}$ and $\mathcal{L}_{KD}$, as shown in Equation \ref{eq:3}:
\begin{equation}\label{eq:3}
    \mathcal{L}_{P} = - \sum_{i=1}^{N} \sum_{j=1}^{m} || \frac{h_{\mathcal{S}}^{i,j}}{||h_{\mathcal{S}}^{i,j}||_2} - \frac{\mathcal{A}(j)^i}{||\mathcal{A}(j)^i||_2}||^2_2
\end{equation}
where $h^{i,j}_\mathcal{S}$ is the output\footnote{By the output, we mean the output of the layer for the {\fontfamily{pcr}\selectfont CLS} token. For more details about {\fontfamily{pcr}\selectfont CLS} see \citet{Devlin2019BERTPO}.} of the $j$-th student layer for the $i$-th input. A subset of teacher layers selected for distillation is denoted with an alignment function $\mathcal{A}$, e.g. $\mathcal{A}(j) = h^{l}_\mathcal{T}$ implies that the output of the $j$-th student layer should be compared to the output of the $l$-th teacher layer ($h_{\mathcal{S}}^{i,j} \leftrightarrow h_{\mathcal{T}}^{i,l}$). 

PKD is not the only model that utilizes internal layers' information. Other models such as TinyBERT \cite{jiao2019tinybert} and MobileBERT \citep{sun2020mobilebert} also found it crucial for training competitive student models. However, as Equation \ref{eq:3} shows, in these models only $m$ teacher layers (the number of teacher layers returned by $\mathcal{A}$) can contribute to distillation. In the presence of deep teachers and small students, this limitation can introduce a significant amount of information loss. Furthermore, what is denoted by $\mathcal{A}$ directly impacts quality. If $\mathcal{A}$ skips an important layer the student model may fail to provide high-quality results. 

To tackle this problem, \citet{wu-etal-2020-skip} proposed a combinatorial technique, called CKD. In their model, $\mathcal{A}(j)$ returns a subset of teacher layers instead of a single layer. Those layers are combined together and distillation happens between the combination result and the $j$-th student layer, as shows in equation \ref{eq:ckd}:
\begin{equation}\label{eq:ckd}
\begin{split}
    \hat{\mathcal{C}}^j =& \mathcal{F}_{c}( h^k_\mathcal{T}) ;\hspace{1mm}{h^k_\mathcal{T} \in \mathcal{A}(j)}
    \\
    \mathcal{C}^j =& \mathcal{F}_{r}(\hat{\mathcal{C}}^j)
    \\
    \cup_{j=1}^m \mathcal{A}(j) =& \{h_\mathcal{T}^1, ..., h_\mathcal{T}^n\}
\end{split}
\end{equation}
where $\hat{\mathcal{C}}^j$ is the result of a combination produced by the function $\mathcal{F}_c$ given a subset of teacher layers indicated by $\mathcal{A}(j)$. In \citet{wu-etal-2020-skip}, $\mathcal{F}_c$ is implemented via a simple concatenation. Depending on the form of combination used in Equation \ref{eq:ckd}, there might be a dimension mismatch between $\hat{\mathcal{C}}^j$ and the student layer $h_{\mathcal{S}}^j$. Accordingly, there is another function, $\mathcal{F}_r$, to reform the combination result into a comparable shape to the student layer. CKD uses a single projection layer to control the dimension mismatch. 

With the combination technique (concatenation+projection), CKD could solve the skip problem but the search problem still remains unanswered. Similar to PKD, CKD also requires a search process, but it looks for the best subset of teacher layers instead of the best single layer. These two models are directly related to this research so we consider them as baselines in our experiments. 

The application of KD in NLP and NLU is not limited to the aforementioned models.  
\citet{aguilar2020knowledge} followed the same architecture as PKD but they introduced a new training regime, called progressive training. In their method, lower layers are trained first and training is progressively shifted to upper layers. They claim that the way internal layers are trained during KD can play a significant role. \citet{Liu2019ImprovingMD} investigated KD from another perspective. Instead of focusing on the compression aspect, they kept the size of student models equal to their teachers and showed how KD could be treated as a complementary training ingredient.

\citet{tan2019multilingual} squeezed multiple translation engines into one transformer \citep{vaswani2017attention} and showed that knowledge can be distilled from multiple teachers. \citet{wei2019online} introduced a novel training procedure where there is no need for an external teacher. A student model can learn from its own checkpoints. At each validation step, if the current checkpoint is better than the best existing checkpoint, student learns from it otherwise the best stored checkpoint is considered as a teacher.

\section{Methodology}
For a given student model $\mathcal{S}$ and a teacher model $\mathcal{T}$ we show all intermediate layers with sets $H_\mathcal{S} = \{h_\mathcal{S}^1, ..., h_\mathcal{S}^m\}$ and $H_\mathcal{T} = \{h_\mathcal{T}^1, ..., h_\mathcal{T}^n\}$, respectively. Based on the pipeline designed by current models for intermediate layer KD, there must be a connection between $H_\mathcal{S}$ and $H_\mathcal{T}$ during training and each student layer can only correspond to a single peer on the teacher side. As previously mentioned, layer connections are denoted by $\mathcal{A}$.  

\begin{figure}[th]
\begin{center}
\includegraphics[scale=0.44]{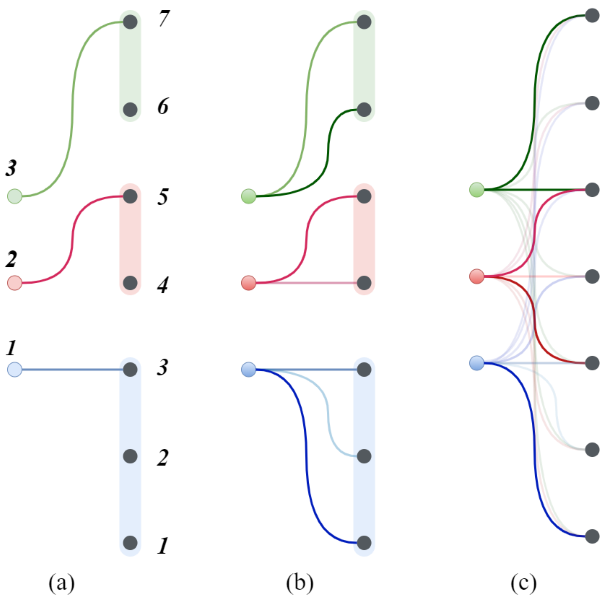}
\end{center}
\caption{\label{fig:1} Three pairs of $\mathcal{S}$ and $\mathcal{T}$ networks with different forms of layer connections. In Figure \ref{fig:1}a, teacher layers are divided into $3$ buckets and only one layer from each bucket is connected to the student side, e.g. $h_\mathcal{T}^5$ is the source of distillation for $h_\mathcal{S}^2$ ($h_\mathcal{T}^5 \leftrightarrow h_\mathcal{S}^2$). In Figure \ref{fig:1}b, a weighted average of teacher layers from each bucket is considered for distillation, e.g. $\mathcal{A}(2)$ = $\{h_\mathcal{T}^4,h_\mathcal{T}^5\}$ and $\mathcal{C}^2=\alpha_{_{24}}h_\mathcal{T}^4 + \alpha_{_{25}}h_\mathcal{T}^5$ ($\mathcal{C}^2 \leftrightarrow h_\mathcal{S}^2$). In Figure \ref{fig:1}c, there is no bucketing and all teacher layers are considered for projection. Links with higher color intensities have higher attention weights.}
\end{figure}

A common heuristic to devise $\mathcal{A}$ is to divide teacher layers into $m$ buckets with approximately the same sizes and pick only one layer from each \citep{jiao2019tinybert,pkd}. Therefore, for the $j$-th layer of the student model, $\mathcal{A}(j)$ returns a single teacher layer among those that reside in the $j$-th bucket. Figure \ref{fig:1}a illustrates this setting. Clearly, this is not the best way of connecting layers, because they are picked in a relatively arbitrary manner. More importantly, no matter what heuristic is used there still remain $n\!-\!m$ layers in this approach whose information is not used in distillation.  

To address this issue, we simply propose a combinatorial alternative whereby all layers inside buckets are taken into consideration. Our technique is formulated in Equation \ref{eq:4}: 
\begin{equation}\label{eq:4}
    \begin{split}
    \mathcal{C}^j =& \sum_{h^k_\mathcal{T} \in \mathcal{A}(j)} \alpha_{jk} \hspace{1mm} h^k_\mathcal{T} 
    \\
    \alpha_{jk} =& \frac{\exp({h^j_\mathcal{S} \hspace{0.5mm} . \hspace{0.5mm}  h^k_\mathcal{T}})}{\sum_{h^{k'}_\mathcal{T} \in \mathcal{A}(j)} \exp({h^j_\mathcal{S} \hspace{0.5mm} . \hspace{0.5mm}h^{k'}_\mathcal{T}})}
    \\
    \cup_{j=1}^m \mathcal{A}(j) &= H_\mathcal{T} = \{h_\mathcal{T}^1, ..., h_\mathcal{T}^n\}
    \end{split}
\end{equation}
This idea is similar to that of CKD, but we use an attention mechanism \cite{bahdanau2014neural} instead of a concatenation for layer combination. Experimental results demonstrate that this form of combination is more useful. We refer to this idea as \textbf{A}ttention-based \textbf{L}ayer \textbf{P}rojection for \textbf{KD} or ALP-KD in short.

According to the equation, if a student layer associates with a particular bucket, all layers inside that bucket are combined/used for distillation and  $\mathcal{C}^j$ is a vector representation of such a combination. Our model benefits from all $n$ teacher layers and skips none as there is a dedicated $\mathcal{C}$ vector for each student layer. Figure \ref{fig:1}b visualizes this setting. 

Weights ($\alpha$ values) assigned to teacher layers are learnable parameters whose values are optimized during training. They show the contribution of each layer to the distillation process. They also reflect the correlation between student and teacher layers, i.e. if a student layer correlates more with a set of teacher layers weights connecting them should receive higher values. In other words, that specific layer is playing the role of its teacher peers on the student side. To measure the correlation, we use the \textit{dot product} in our experiments but any other function for similarity estimation could be used in this regard. 

Equation \ref{eq:4} addresses the \textit{skip} problem with a better combination mechanism and is able to provide state-of-the-art results. However, it still suffers from the \textit{search} problem as it relies on buckets and we are not sure which bucketing strategy works better. For example, in Figure \ref{fig:1}b the first bucket consists of the first three layers of the teacher but it does not mean that we cannot append a fourth layer. In fact, a bucket with four layers might perform better. Buckets can also share layers; namely, a teacher layer can belong to multiple buckets and can be used numerous times in distillation. These constraints make it challenging to decide about buckets and their boundaries, but it is possible to resolve this dilemma through a simple modification in our proposed model. 

To avoid bucketing, we span the attention mask over all teacher layers rather than over buckets. To implement this extension, $\mathcal{A}(j)$ needs to be replaced with $H_\mathcal{T}$ in Equation \ref{eq:4}. Therefore, for any student layer such as $h_\mathcal{S}^j$ there would be a unique set of $n$ attention weights and $\mathcal{C}^j$ would be a weighted average of \textit{all} teacher layers, as shown in Equation \ref{eq:5}: 
\begin{equation}\label{eq:5}
    \begin{split}
\mathcal{C}^j =& \sum_{h^k_\mathcal{T} \in \mathcal{A}(j)} \alpha_{jk} \hspace{1mm} h^k_\mathcal{T}\\
\mathcal{A}(j) =& H_\mathcal{T} \hspace{2mm} \forall j\in \{1,2,...,m\}
\end{split}
\end{equation}
This new configuration, which is illustrated in Figure \ref{fig:1}c, proposes a straightforward way of combining teacher layers and addresses both \textit{skip} and \textit{search} problems at the same time. 

To train our student models, we use a loss function which is composed of $\mathcal{L}_{ce}$, $\mathcal{L}_{KD}$, and a dedicated loss defined for ALP-KD, as shown in Equation \ref{eq:6}: 
\begin{equation}\label{eq:6}
    \begin{split}
    \mathcal{L} = \beta \mathcal{L}_{ce} + \eta \mathcal{L}_{KD} + \lambda \mathcal{L}_{_{ALP}}
    \\
    \mathcal{L}_{_{ALP}} = \sum_{i=1}^{N}\sum_{j=1}^{m} \textrm{\textit{MSE}}(h_{\mathcal{S}}^{i,j},\mathcal{C}^{i,j})
    \end{split}
\end{equation}
where \textit{MSE()} is the mean-square error and $\mathcal{C}^{i,j}$ shows the value of $\mathcal{C}^j$ when the teacher is fed with the $i$-th input. $\beta$, $\eta$, and $\lambda$ are hyper-parameters of our model to minimize the final loss. 

\section{Experimental Study}\label{exp} 
A common practice in our field to evaluate the quality of a KD technique is to feed $\mathcal{T}$ and $\mathcal{S}$ models with instances of standard datasets and measure how they perform. We followed the same tradition in this paper and selected a set of eight GLUE tasks \cite{Wang2018GLUEAM} including CoLA, MNLI, MRPC, QNLI, QQP, RTE, SST-2, and STS-B datasets to benchmark our models. Detailed information about datasets is available in the appendix section. 

\begin{table*}[hbt!]
\centering
\begin{tabular}{l l l l l l l l l l l}
\toprule
Problem &Model & CoLA & MNLI & MRPC & QNLI & QQP & RTE & SST-2 & STS-B & Average\\
\hline
N/A & $\mathcal{T}_{_{\textrm{BERT}}}$ & 57.31 & 83.39 & 86.76 &	91.25 &	90.96	& 68.23	& 92.67	& 88.82	& 82.42\\\hline
N/A & $\mathcal{S}_{_{\textrm{NKD}}}$ & 31.05 &	76.83	& 77.70 & 85.13	& 88.97	& 61.73 &	88.19	& 87.29	& 74.61\\
\textit{skip}, \textit{search} &	 $\mathcal{S}_{_{\textrm{RKD}}}$	&29.22	&79.31	&79.41	&86.77	&90.25	&65.34	&90.37	&87.45	&76.02
\\\hline
\textit{skip}, \textit{search} &	 $\mathcal{S}_{_{\textrm{PKD}}}$	&32.13	&79.26	&80.15	&86.64	&90.23	&65.70	&90.14	&87.26	&76.44
\\\hline
\textit{search}	&  $\mathcal{S}_{_{\textrm{CKD-NO}}}$ &	31.23 & 79.42	&80.64	&86.93	&88.70	&66.06	&90.37	&87.62	&76.37
\\
\textit{search}	&  $\mathcal{S}_{_{\textrm{CKD-PO}}}$ &31.95& 79.53&	80.39	&86.75&	89.89&	\textbf{67.51}&	90.25&	87.55&	76.73
\\\hline
\textit{search} &	 $\mathcal{S}_{_{\textrm{ALP-NO}}}$ &	\textbf{34.21}&	79.26	&79.66	&\textbf{87.11}&	\textbf{90.72}&	65.70&	90.37&	87.52&	76.82
\\
\textit{search} &	 $\mathcal{S}_{_{\textrm{ALP-PO}}}$ &	33.86&	\textbf{79.74}&	79.90&	86.95&	90.25&	66.43	&\textbf{90.48}	&87.52&	76.89
\\\hline
\textit{none}   & $\mathcal{S}_{_{\textrm{ALP}}}$&	33.07&	{79.62}&	\textbf{80.72}&	87.02&	90.54&	67.15&	90.37&	\textbf{87.62}&	\textbf{77.01}
\\
\bottomrule
\end{tabular}
\caption{\label{t:1} Except the teacher ($\mathcal{T}_{_{\textrm{BERT}}}$) which is a $12$-layer model, all other models have $4$ layers. Apart from the number of layers, all students have the same architecture as the teacher. The first column shows what sort of problems each model suffers from. NKD stands for \textit{No KD} which means there is no KD technique involved during training this student model. \textit{NO} and \textit{PO} are different configurations for mapping internal layers. Boldfaced numbers show the best student score for each column over the validation set. Scores in the first column are Matthew's Correlations. SST-B scores are Pearson correlations and the rest are accuracy scores.}
\end{table*}

In NLP/NLU settings, $\mathcal{T}$ is usually a pre-trained model whose parameters are only fine-tuned during training. On the other side, $\mathcal{S}$ can be connected to $\mathcal{T}$ to be trained thoroughly or can alternatively be initialized with $\mathcal{T}$'s parameters to be fine-tuned similar to its teacher. This helps the student network generate better results and converge faster. {Fine-tuning} is more common than {training} in our context and we thus fine-tune our models rather than training. This concept is comprehensively discussed by \citet{Devlin2019BERTPO} so we skip its details and refer the reader to their paper. We have the same fine-tuning pipeline in this work. 

In our experiments, we chose the original BERT model\footnote{\url{https://github.com/google-research/bert}} (also known as BERT$_{\textrm{Base}}$) as our teacher. We are faithful to the configuration proposed by \citet{Devlin2019BERTPO} for it. Therefore, our in-house version also has $12$ layers with $12$ attention heads and the hidden and feed-forward dimensions are $768$ and $3072$, respectively. Our students are also BERT models only with fewer layers ($|H_\mathcal{S}| = m\hspace{1mm};m<12$). We use the teacher BERT to initialize students, but because the number of layers are different ($12 \ne m$) we only consider its first $m$ layers. We borrowed this idea from PKD \cite{pkd} in the interest of fair comparisons.

In order to maximize each student's performance we need to decide about the learning rate, batch size, the number of fine-tuning iterations, and $\beta$, $\eta$, and $\lambda$. To this end, we run a grid search similar to \citet{pkd} and \citet{wu-etal-2020-skip}. In our setting, the batch size is set to $32$ and the learning rate is selected from $\{1e-5,2e-5, 5e-5\}$. $\eta$ and $\lambda$ take values from $\{0, 0.2, 0.5, 0.7\}$ and $\beta = 1-\eta-\lambda$. Details of the grid search and values of all hyper-parameter are reported in the appendix section.  

We trained multiple models with different configurations and compared our results to RKD- and PKD-based students. To the best of our knowledge, these are the only alternatives that use BERT as a teacher and their students' architecture relies on ordinary Transformer blocks \citep{vaswani2017attention} with the same size as ours, so any comparison to any other model with different settings would not be fair. Due to CKD's similarity to our approach we also re-implemented it in our experiments. The original CKD model was proposed for machine translation and for the first time we evaluate it in NLU tasks. Table \ref{t:1} summarizes our experiments. 

The teacher model with $12$ layers and $109$M parameters has the best performance for all datasets.\footnote{Similar to other papers, we evaluate our models on validation sets. Testset labels of GLUE datasets are not publicly available and researchers need to participate in leaderboard competitions to evaluate their models on testsets.} This model can be compressed, so we reduce the number of layers to $4$ and train another model ($\mathcal{S}_{_{\textrm{NKD}}}$). The rest of the configuration (attention heads, hidden dimension etc) remains untouched. There is no connection between the teacher and $\mathcal{S}_{_{\textrm{NKD}}}$ and it is trained separately with no KD technique. Because of the number of layers, performance drops in this case but we still gain a lot in terms of memory as this new model only has $53$M parameters. To bridge the performance gap between the teacher and $\mathcal{S}_{_{\textrm{NKD}}}$, we involve KD in the training process and train new models, $\mathcal{S}_{_{\textrm{RKD}}}$ and $\mathcal{S}_{_{\textrm{PKD}}}$, with RKD and PKD techniques, respectively.   

$\mathcal{S}_{_{\textrm{RKD}}}$ is equivalent to a configuration known as DistilBERT in the literature \cite{Sanh2019DistilBERTAD}. To have precise results and a better comparison, we trained/fine-tuned all models in the same experimental environment. Accordingly, we do not borrow any result from the literature but reproduce them. This is the reason we use the term equivalent for these two models. Furthermore, DistilBERT has an extra Cosine embedding loss in addition to those of $\mathcal{S}_{_{\textrm{RKD}}}$. When investigating the impact of intermediate layers in the context of KD, we wanted $\mathcal{L}_{P}$ to be the only difference between RKD and PKD, so incorporating any other factor could hurt our investigation and we thus avoided the cosine embedding loss in our implementation. 

PKD outperforms RKD with an acceptable margin in Table \ref{t:1} and that is because of the engagement of intermediate layers. For $\mathcal{S}_{_{\textrm{PKD}}}$, we divided teacher layers into $3$ buckets ($4$ layers in each) and picked the first layer of each bucket to connect to student layers, i.e. $\mathcal{A}(1)=h_{\mathcal{T}}^1$, $\mathcal{A}(2)=h_{\mathcal{T}}^5$, and $\mathcal{A}(3)=h_{\mathcal{T}}^9$. There is no teacher layer assigned to the last layer of the student. This form of mapping maximizes PKD's performance and we figured out this via an empirical study. 

\begin{table*}[ht!]
\centering
\begin{tabular}{l l l l l l l l l l l}
\toprule
Problem &Model & CoLA & MNLI & MRPC & QNLI & QQP & RTE & SST-2 & STS-B & Average\\
\hline
N/A & $\mathcal{T}_{_{\textrm{BERT}}}$ & 57.31&	83.39&	86.76&	91.25&	90.96&	68.23&	92.67&	88.82&	82.42

\\\hline
N/A & $\mathcal{S}_{_{\textrm{NKD}}}$ & 40.33&	79.91&	81.86&	87.57	&90.21	&65.34	&90.02&	88.49&	77.97

\\
\textit{skip}, \textit{search} & $\mathcal{S}_{_{\textrm{RKD}}}$ & 45.51	&81.41&	83.82&	88.21&	90.56&	67.51&	91.51&	88.70&	79.65

\\\hline
\textit{skip}, \textit{search} & $\mathcal{S}_{_{\textrm{PKD}}}$ & 45.78&	\textbf{82.18}&	85.05&	89.31&	90.73&	68.23&	91.51&	88.56&	80.17

\\\hline
\textit{search}	&  $\mathcal{S}_{_{\textrm{CKD-NO}}}$ & \textbf{48.49}&	81.91&	83.82&	89.53&	90.64&	67.51&	91.40&	88.73&	80.25

\\
\textit{search}	&  $\mathcal{S}_{_{\textrm{CKD-PO}}}$ & 46.99	&81.99&	83.82&	89.44	&\textbf{90.82}&	67.51&	91.17&	88.62&	80.05

\\\hline
\textit{search} &	 $\mathcal{S}_{_{\textrm{ALP-NO}}}$ & 46.40&	81.99&\textbf{85.78}&	\textbf{89.71}&	90.64&	\textbf{68.95}&	\textbf{91.86}&	\textbf{88.81}&	\textbf{80.52}

\\
\textit{search} &	 $\mathcal{S}_{_{\textrm{ALP-PO}}}$ & 46.02&	82.04	&84.07&	89.16&	90.56&	68.23&	91.74&	88.72	&80.07

\\\hline
\textit{none}   & $\mathcal{S}_{_{\textrm{ALP}}}$& 46.81&	81.86&	85.05&	89.67&	90.73&	68.59&	\textbf{91.86}&	88.68&	80.41
\\
\bottomrule
\end{tabular}
\caption{\label{t:2} The teacher model $\mathcal{T}_{_{\textrm{BERT}}}$ has $12$ and all other student models have $6$ layers.}
\end{table*}

Results discussed so far demonstrate that cross-model layer mapping is effective, but it can be improved even more if the skip issue is settled. Therefore, we trained two other students using CKD. The setting for these models is identical to PKD, namely teacher layers are divided into $3$ buckets. The first $4$ teacher layers reside in the first bucket. The fifth to eighth layers are in the second bucket and the rest are covered by the third bucket. Layers inside the first bucket are concatenated and passed through a projection layer to match the student layers' dimension. The combination result for the first bucket is assigned to the first student layer ($\mathcal{C}^1 \leftrightarrow h_{\mathcal{S}}^1$). The same procedure is repeated with the second and third buckets for $h_{\mathcal{S}}^2$ and $h_{\mathcal{S}}^3$. Similar to PKD, there is no teacher layer connected to the last student layer. This configuration is referred to as \textbf{N}o \textbf{O}verlap (\textbf{NO}), that indicates  buckets share no layers with each other. 

In addition to \textbf{NO} we designed a second configuration, \textbf{PO}, which stands for \textbf{P}artial \textbf{O}verlap. In \textbf{PO}, each bucket shares its first layer with the preceding bucket, so the first bucket includes the first to fifth layers, the second bucket includes the fifth to ninth layers, and from the ninth layer onward reside in the third bucket. We explored this additional configuration to see the impact of different bucketing strategies in CKD. %The fifth and sixth rows in Table \ref{t:1} summarize results of \textbf{NO} and \textbf{PO}. 

Comparing $\mathcal{S}_{_{\textrm{CKD}}}$ to  $\mathcal{S}_{_{\textrm{PKD}}}$ shows that the combination (concatenation+projection) idea is useful in some cases, but for others the simple skip idea is still better. Even defining different bucketing strategies did not change it drastically, and this leads us to believe that a better form of combination such as an attention-based model is required.

In $\mathcal{S}_{_{\textrm{ALP}}}$ extensions, we replace the CKD's concatenation with attention and results improve. ALP-KD is consistently better than all other RKD, PKD, and CKD variations and this justifies the necessity of using attention for combination. $\mathcal{S}_{_{\textrm{ALP-NO}}}$ and $\mathcal{S}_{_{\textrm{ALP-PO}}}$ also directly support this claim. In $\mathcal{S}_{_{\textrm{ALP}}}$, we followed Equation \ref{eq:5} and spanned the attention mask over all teacher layers. This setting provides a model that requires no engineering adjustment to deal with \textit{skip} and \textit{search} problems and yet delivers the best result on average.

\subsection{Training Deeper/Shallower Models Than $4$-Layer Students}
So far we compared $4$-layer ALP-KD models to others and observed superior results. In this section, we design additional experiments to study our technique's behaviour from the size perspective. The original idea of PKD was proposed to distill from a $12$-layer BERT to a $6$-layer student \cite{pkd}. In such a scenario, only every other layer of the teacher is skipped and it seems the student model should not suffer from the skip problem dramatically. We repeated this experiment to understand if our combination idea is still useful or its impact diminishes when student and teacher models have closer architectures. Table \ref{t:2} summarizes findings of this experiment. 

Among $6$-layer students, $\mathcal{S}_{_{\textrm{ALP-NO}}}$ has the best average score which demonstrates that the combinatorial approach is still useful. Moreover, the supremacy of attention-based combination over the simple concatenation holds for this setting too. $\mathcal{S}_{_{\textrm{ALP}}}$ is the second best and yet our favorite model as it requires no layer alignment before training.

The gap between PKD and ALP-KD is narrowed in $6$-layer models compared to $4$-layer students, and this might be due to an implicit relation between the size and need for combining intermediate layers. We focused on this hypothesis in another experiment and this time used the same teacher to train $2$-layer students. In this scenario, student models are considerably smaller with only $39$M parameters. Results of this experiment are reported in Table \ref{t:3}.

\begin{table*}[ht!]
\centering
\begin{tabular}{l l l l l l l l l l l}
\toprule
Problem &Model & CoLA & MNLI & MRPC & QNLI & QQP & RTE & SST-2 & STS-B & Average\\
\hline
N/A & $\mathcal{T}_{_{\textrm{BERT}}}$ & 57.31&	83.39&	86.76&	91.25&	90.96&	68.23&	92.67&	88.82&	82.42

\\\hline
N/A & $\mathcal{S}_{_{\textrm{NKD}}}$ & 14.50&	72.73&	72.06&	79.61	&86.89&	57.04&	85.89&	40.80&	63.69
\\
\textit{skip}, \textit{search} & $\mathcal{S}_{_{\textrm{RKD}}}$ & 24.50&	74.90&	73.53&	81.04&	87.40&	59.21	&87.39&	41.87&	66.23

\\\hline
\textit{skip}, \textit{search} & $\mathcal{S}_{_{\textrm{PKD-1}}}$ & 23.09&	74.65&	72.55&	81.27	&87.68&	57.40&	88.76&	43.37&	66.1

\\
\textit{skip}, \textit{search}	&  $\mathcal{S}_{_{\textrm{PKD-6}}}$ &22.48&	74.57&	73.04&	80.74&	87.70&	57.40&	88.65&	42.92&	65.94

\\
\textit{skip}, \textit{search}	&  $\mathcal{S}_{_{\textrm{PKD-12}}}$ & 22.46	&74.33&	72.79&	81.22&	87.88&	57.40&	88.76&	45.39&	66.28
\\\hline
\textit{search}	&  $\mathcal{S}_{_{\textrm{CKD}}}$ & \textbf{24.69}&	74.67&	73.04&	\textbf{81.60}&	87.10	&58.84&	88.65&	43.71&	66.54

\\\hline
\textit{none}   & $\mathcal{S}_{_{\textrm{ALP}}}$& 24.61	&\textbf{74.78}&	\textbf{73.53}&	81.24&	\textbf{88.01}	&\textbf{59.57}&	\textbf{88.88}&	\textbf{46.04}&	\textbf{67.08}
\\
\bottomrule
\end{tabular}
\caption{\label{t:3} The teacher model $\mathcal{T}_{_{\textrm{BERT}}}$ has $12$ and all other student models have $2$ layers. $\mathcal{S}_{_{\textrm{PKD-\textit{l}}}}$ indicates that $h_\mathcal{T}^l$ is used for distillation.}
\end{table*}

For CKD and ALP-KD, we combine all teacher layers and distill into the first layer of the student. Similar to previous experiments, there is no connection between the last layer of $2$-layer students and the teacher model and KD happens between $h_\mathcal{S}^1$ and $H_\mathcal{T}$. For PKD, we need to decide which teacher layers should be involved in distillation, for which we assessed three configurations with the first ($h_\mathcal{S}^1 \leftrightarrow h_\mathcal{T}^1$), sixth ($h_\mathcal{S}^1 \leftrightarrow h_\mathcal{T}^6$), and twelfth ($h_\mathcal{S}^1 \leftrightarrow h_\mathcal{T}^{12}$) layers. $\mathcal{S}_{_{\textrm{ALP}}}$ outperforms other students in this case too and this time the gap between PKD and ALP-KD is even more visible. This result points out to the fact that when teacher and student models differ significantly, intermediate layer combination becomes crucial. 

\subsection{Qualitative Analysis}
\begin{figure}[th]
%\begin{flushleft}
\includegraphics[scale=0.23]{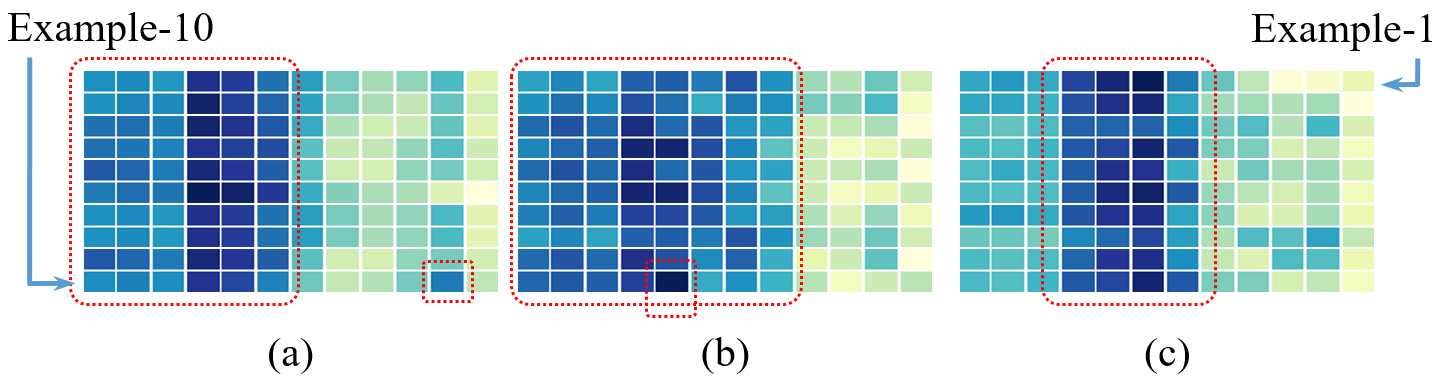}
%\end{flushleft}
\caption{\label{fig:3} Visualizing attention weights between the first layer of the student model and all teacher layers for $10$ samples from SST-2. Weights belong to $\mathcal{S}_{_{\textrm{ALP}}}$ with $2$ (a),  $4$ (b), and $6$ (c) layers.}
\end{figure}

We tried to visualize attention weights to understand what happens during training and why ALP-KD leads to better performance.  Figure \ref{fig:3} illustrates results related to this experiment. From the SST-2 dataset, we randomly selected $10$ examples and stimulated both teacher and student models to emit attention weights between the first layer of the student ($h_\mathcal{S}^1$) and all teacher layers ($H_\mathcal{T}$). We carried out this experiment with $2$-, $4$-, and $6$-layer $\mathcal{S}_{_{\textrm{ALP}}}$ models. The \textit{x} and \textit{y} axes in the figure show the attention weights and $10$ examples, respectively. 

As seen in Figure \ref{fig:3}a, the first half of the teacher model is more active, which is expected since we distill into the first layer of the student. However, $h_\mathcal{S}^{1}$ receives strong signals from other layers in the second half too, e.g. in {\fontfamily{pcr}\selectfont Example-10} there is a strong connection between $h_\mathcal{T}^{11}$ and $h_\mathcal{S}^{1}$. This visualization demonstrates that all teacher layers participate in distillation and defining buckets or skipping layers might not be the best approach. A similar situation arises when distilling into the $4$-layer model in Figure \ref{fig:3}b as the first half is still more active. For the $6$-layer model, we see a different pattern where there is a concentration in attention weights around the middle layers of the teacher and $h_\mathcal{S}^1$ is mainly fed by layers $h_\mathcal{T}^4$ to $h_\mathcal{T}^7$.

Considering the distribution of attention weights, any skip- or even concatenation-based approach would fail to reveal the maximum capacity of KD. Such approaches assume that a single teacher layer or a subset of adjacent layers affect the student model, whereas almost all of them participate in the process. Apart from previously reported results, this visualization again justifies the need for an attention-based combination in KD. 

\begin{figure}
\centering
\begin{subfigure}{.25\textwidth}
  \centering
  \includegraphics[width=.92\linewidth]{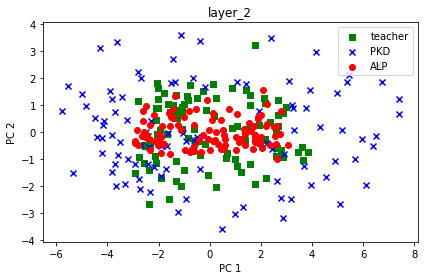}
  \caption{}
  \label{fig:sub1}
\end{subfigure}%
\begin{subfigure}{.25\textwidth}
  \centering
  \includegraphics[width=.92\linewidth]{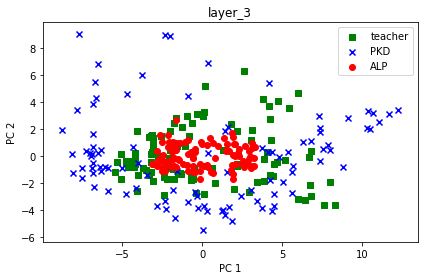}
  \caption{}
  \label{fig:sub2}
\end{subfigure}
\begin{subfigure}{.25\textwidth}
  \centering
  \includegraphics[width=.93\linewidth]{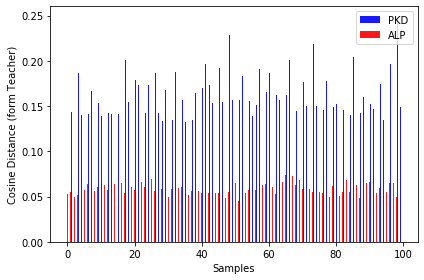}
  \caption{}
  \label{fig:sub3}
\end{subfigure}%
\begin{subfigure}{.25\textwidth}
  \centering
  \includegraphics[width=.93\linewidth]{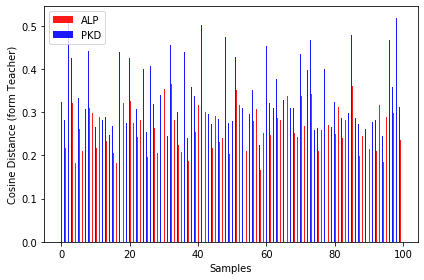}
  \caption{}
  \label{fig:sub4}
\end{subfigure}
\caption{Visualizing intermediate layers' outputs and their distance from the teacher in ALP-KD and PKD students. Teacher-, ALP-KD-, and PKD-related information is visualized with {green}, {red}, and {blue} colors, respectively. Figures 4a and 4c provide information about $h^{2}_{\textrm{ALP}}$, $h^{2}_{\textrm{PKD}}$, and $h_\mathcal{T}^5$, and Figures 4b and 4d report information about $h^{3}_{\textrm{ALP}}$, $h^{3}_{\textrm{PKD}}$, and $h_\mathcal{T}^9$. In the bottom figures, the \textit{x} axis shows samples and the \textit{y} axis is the Cosine distance from the teacher.}
\label{fig:internal-layers}
\end{figure}

Our technique emphasizes on intermediate layers and the necessity of having similar internal representations between student and teacher models, so in addition to attention weights we also visualized the output of intermediate layers. The main idea behind this analysis is to show the information flow inside student models and how ALP-KD helps them mimic their teacher. Figures \ref{fig:internal-layers}a and \ref{fig:internal-layers}b illustrate this experiment.

We randomly selected $100$ samples from the SST-2 dataset and visualized what hidden representations of  $\mathcal{S}_{_{\textrm{ALP}}}$, $\mathcal{S}_{_{\textrm{PKD}}}$, and $\mathcal{T}$ models (from Table \ref{t:1}) look like when stimulated with these inputs. Student models have $4$ layers but due to space limitations we only show middle layers' outputs, namely $h_\mathcal{S}^2$ (Figure \ref{fig:internal-layers}a) and $h_\mathcal{S}^3$ (Figure \ref{fig:internal-layers}b). $h_\mathcal{S}^1$ and $h_\mathcal{S}^4$ also expressed very similar attitudes. 

The output of each intermediate layer is a $768$-dimensional vector, but for visualization purposes we consider the first two principle components extracted via PCA \cite{wold1987principal}. During training, $h_\mathcal{T}^5$ and $h_\mathcal{T}^9$ are connected to $h_\mathcal{S}^2$ and $h_\mathcal{S}^3$ as the source of distillation in PKD, so we also include those teacher layers' outputs in our visualization. As the figure shows, ALP-KD's representations are closer to teacher's and it demonstrates that our technique helps train better students with closer characteristics to teachers.

We conducted another complementary analysis where we used the output of the same teacher and student layers from the previous experiment and measured their distance for all $100$ examples. Results of this experiment are illustrated in Figures \ref{fig:internal-layers}c and \ref{fig:internal-layers}d for the second and third student layers, respectively. Internal representations generated by PKD are more distant from those of the teacher compared to ALP-KD's representations, e.g. the distance between $h^{20,2}_{\textrm{PKD}}$ (the output of the second PKD layer for the $20$-\textit{th} example in Figure \ref{fig:internal-layers}c) and $h^{20,5}_{\mathcal{T}}$ is around $0.20$ whereas this number is only $0.05$ for ALP-KD. This is an indication that the ALP-KD student follows its teacher better than the PKD student. To measure distance, we used the Cosine similarity in this experiment.  

\section{Conclusion and Future Work}
In this paper, we discussed the importance of distilling from intermediate layers and proposed an attention-based technique to combine teacher layers without skipping them. Experimental results show that the combination idea is effective. Our findings in this research can be summarized as follows:
\begin{itemize}
    \item It seems to distil from deep teachers with multiple internal components combination is essential.
    
    \item The more teacher and student models differ in terms of the number of layers, the more intermediate layer combination becomes crucial.
    
    \item Although a simple concatenation of layers is still better than skipping in many cases, to obtain competitive results an attention-based combination is required.
    
    \item ALP-KD can be tuned to combine layers inside buckets and this approach is likely to yield state-of-the-art results, but if there is no enough knowledge to decide about buckets, a simple attention mask over all teacher layers should solve the problem. 
\end{itemize}

As our future direction, we are interested in applying ALP-KD to other tasks to distill from extremely deep teachers into compact students. Moreover, we will work on designing better attention modules. Techniques that are able to handle sparse structures could be more useful in our architecture. Finally, we like to adapt our model to combine other internal components such as attention heads. 

\section*{Acknowledgement}
We would like to thank our anonymous reviewers as well as Chao Xing and David Alfonso Hermelo from Huawei Noah's Ark Lab for their valuable feedback.

\bibliography{aaai2021}

\clearpage
\section*{Appendix}
\subsection*{GLUE Datasets}
Datasets used in our experiments are as follows: 
\begin{itemize}
    \item \textbf{CoLA}: A corpus of English sentences drawn from books and journal articles with $8,551$ training and $1,043$ validation instances. Each example is a sequence of words with a label indicating whether it is a grammatical sentence \cite{warstadt2018neural}. 
    
    \item \textbf{MNLI}: A multi-genre natural language inference corpus including sentence pairs with textual entailment annotations \cite{N18-1101}. The task defined based on this dataset is to predict  whether the premise entails the hypothesis, contradicts it, or neither, given a premise sentence and a hypothesis. The dataset has two versions, \textit{matched} (test and training examples are from the same domain) and \textit{mismatched}, that we use the matched version. This dataset has $392,702$ training and $9,815$ validation examples.
    
    \item \textbf{MRPC}: A corpus of sentence pairs with human annotations. The task is to decide whether sentences are semantically equivalent \cite{dolan2005automatically}. The training and validation sets have $3,668$ and $408$ examples, respectively.
    
    \item \textbf{QNLI}: A dataset built for a binary classification task to assess whether a sentence contains the correct answer to a given query \cite{rajpurkar2016squad}. The set has $104,743$ training and $5,463$ validation examples. 
    
    \item \textbf{QQP}: A set of question pairs with $363,849$ training and $40,430$ validation instances collected from the well-known question answering website Quora. The task is to determine if a given pair of questions are semantically equivalent \cite{WinNT}.
    
    \item \textbf{RTE}: A combined set of $2,490$ training and $277$ validations examples collected from four sources for a series of textual entailment challenges \cite{dagan2005pascal,bar2006second,giampiccolo2007third,bentivogli2009fifth}.  
    
    \item \textbf{SST-2}: A sentiment analysis dataset with sentence-level (positive/negative) labels. The training and validation sets include $67,349$ and $872$ sentences, respectively \cite{socher2013recursive}.
    
    \item \textbf{STS-B}: A collection of sentence pairs used for semantic similarity estimation. Each pair has a similarity score from $1$ to $5$. This dataset has $5,749$ training and $1,500$ validation examples \cite{cer2017semeval}.

\end{itemize}

\subsection*{Different Attention Models}
The attention mechanism is the main reason that our architecture works better (than PKD and CKD). The default module designed in ALP-KD relies on a simple dot product, but we studied if a better attention technique can boost performance even more. Accordingly, we adapted the idea of \citet{vaswani2017attention} and carried out a new experiment, which is reported in Table \ref{t:4}. 

To measure the correlation between $h_\mathcal{S}^j$ and teacher layers, we consider $h_\mathcal{S}^j$ as a \textit{query} vector, teacher layers as \textit{key} vectors, and train a dedicated \textit{value} vector for each key. We compute attention weights using key, query, and value vectors as described in \citet{vaswani2017attention}. For this extension, we implemented single- and multi-head attention modules. 

According to Table \ref{t:4}, as the number of heads increases performance improves accordingly, but neither the single-head nor the $4$-head model could outperform the simple dot-product-based technique, which was unexpected.   

\begin{table*}[ht!]
\centering
\begin{tabular}{l l l l l l l l l l l}
\toprule
Model & CoLA & MNLI & MRPC & QNLI & QQP & RTE & SST-2 & STS-B & Average\\
\hline
$\mathcal{S}_{_{\textrm{ALP-NO-1}}}$& 32.84	&78.94&	79.03&	86.31&	90.31&	65.06	&88.53&	87.26&	76.04

\\
$\mathcal{S}_{_{\textrm{ALP-PO-1}}}$& 31.56&	78.37&	79.66&	85.39&	89.97&	65.32&	89.04&	87.45&	75.85

\\
$\mathcal{S}_{_{\textrm{ALP-1}}}$& 31.75&	79.31&	\textbf{80.64}&	86.75&	90.11&	66.06&	89.95&	87.33&	76.49

\\\hline
$\mathcal{S}_{_{\textrm{ALP-NO-4}}}$& 33.37&
\textbf{79.62}&\textbf{80.64}&	86.84&	\textbf{90.27}&	66.06&	89.91&	\textbf{87.67}&	76.8

\\
$\mathcal{S}_{_{\textrm{ALP-PO-4}}}$& \textbf{33.62}&	79.50&	79.90&	86.69&	90.14&	65.34&	90.25&	87.52&	76.62

\\
$\mathcal{S}_{_{\textrm{ALP-4}}}$& 32.61&	79.34&	79.90&	\textbf{87.00}&	90.26&	\textbf{67.31}&	\textbf{90.71}&	87.46&	\textbf{76.82}
\\
\bottomrule
\end{tabular}
\caption{\label{t:4} The impact of different attention techniques. All models use the key-query-value architecture to compute weights. Digits appended to subscripts indicate the number of attention heads.}
\end{table*}

\subsection*{Hyper-parameters}
We set the batch size to $32$ for all models. The maximum sequence length is $64$ for single-sentence tasks and $128$ for other sentence-pair tasks. For STS-B, SST-2, and QNLI we run $10$ epochs for fine-tuning. For datastes with more than $300$K training instances we only run $5$ epochs. For MRPC and RTE we fine-tune the model for $20$ epochs, and CoLA is the only dataset which needs $50$ epochs to provide high-quality results. For these numbers we tried to follow the literature and have the same experimental settings as others. 

We first initialize student models and then fine-tune both $\mathcal{T}_{_{\textrm{BERT}}}$ and $\mathcal{S}_{_{\textrm{NKD}}}$ with a learning rate chosen from $\{1e-5, 2e-5, 5e-5\}$. For $\mathcal{S}_{_{\textrm{RKD}}}$, $\eta$ takes values from $\{0.2, 0.5, 0.7\}$ and $\beta$ is set to $1-\eta$. $\lambda$ is $0$ since there is no intermediate-layer distillation. For the Softmax function we consider $1$, $5$, $10$, or $20$ as potential values for the temperature (T) \cite{hintonkd}. For each task, a grid search is performed over the learning-rate set, $\eta$, and T. 

For $\mathcal{S}_{_{\textrm{PKD}}}$, the process is almost the same with a single difference. There is an additional hyper-parameter $\lambda$, which incorporates the effect of the intermediate-layer loss. It takes values from $\{0.2, 0.5, 0.7\}$. Unlike the previous setting, this time $\beta$ is $1-\eta-\lambda$. In $\mathcal{S}_{_{\textrm{PKD}}}$, we conduct a grid search over $\lambda$ as well as all other hyper-parameters. For $\mathcal{S}_{_{\textrm{CKD}}}$ and $\mathcal{S}_{_{\textrm{ALP}}}$ models we use the same grid search as in $\mathcal{S}_{_{\textrm{PKD}}}$. Tables \ref{t:5} to \ref{t:10} list the exact values of hyper-parameters for all experiments. 

\begin{table*}[ht!]
\centering
\begin{tabular}{l l l l l l l l l l}
\toprule
hyper-parameters & CoLA & MNLI & MRPC & QNLI & QQP & RTE & SST-2 & STS-B \\
\hline
learning rate& 2e-5	&5e-5&	2e-5&	2e-5&	2e-5&	2e-5	&2e-5&	2e-5
\\
\bottomrule
\end{tabular}
\caption{\label{t:5} Hyper-parameters of $\mathcal{T}_{_{\textrm{BERT}}}$.}
\end{table*}

\begin{table*}[ht!]
\centering
\begin{tabular}{l l l l l l l l l l}
\toprule
hyper-parameters & CoLA & MNLI & MRPC & QNLI & QQP & RTE & SST-2 & STS-B \\
\hline
learning rate& 2e-5	&5e-5&	2e-5&	2e-5&	5e-5&	2e-5	&5e-5&	5e-5
\\
\bottomrule
\end{tabular}
\caption{\label{t:6} Hyper-parameters of $\mathcal{S}_{_{\textrm{NKD}}}$.}
\end{table*}

\begin{table*}[ht!]
\centering
\begin{tabular}{l l l l l l l l l l}
\toprule
hyper-parameters & CoLA & MNLI & MRPC & QNLI & QQP & RTE & SST-2 & STS-B \\
\hline
learning rate& 2e-5	&5e-5&	2e-5&	5e-5&	5e-5&	5e-5	&5e-5&	5e-5
\\
T& 10& 20& 10& 20& 20& 5& 10& 5
\\
$\eta$& 0.5& 0.7& 0.5& 0.7& 0.7& 0.7& 0.7& 0.2
\\
\bottomrule
\end{tabular}
\caption{\label{t:7} Hyper-parameters of $\mathcal{S}_{_{\textrm{RKD}}}$.}
\end{table*}

\begin{table*}[ht!]
\centering
\begin{tabular}{l l l l l l l l l l}
\toprule
hyper-parameters & CoLA & MNLI & MRPC & QNLI & QQP & RTE & SST-2 & STS-B \\
\hline
learning rate& 2e-5	&5e-5&	5e-5&	5e-5&	5e-5&	2e-5	&5e-5&	5e-5
\\
T& 5& 20& 5& 20& 20& 5& 5& 5
\\
$\eta$& 0.5& 0.2& 0.2& 0.2& 0.5& 0.2& 0.2& 0.2
\\
$\lambda$ & 0.2& 0.7& 0.7& 0.7& 0.2& 0.7& 0.7& 0.2
\\
\bottomrule
\end{tabular}
\caption{\label{t:8} Hyper-parameters of $\mathcal{S}_{_{\textrm{PKD}}}$.}
\end{table*}

\begin{table*}[ht!]
\centering
\begin{tabular}{l l l l l l l l l l}
\toprule
hyper-parameters & CoLA & MNLI & MRPC & QNLI & QQP & RTE & SST-2 & STS-B \\
\hline
learning rate& 5e-5	&5e-5&	2e-5&	5e-5&	5e-5&	5e-5	&5e-5&	5e-5
\\
T& 20& 10& 10& 20& 5& 10& 20& 20
\\
$\eta$& 0.2& 0.5& 0.7& 0.7& 0.2& 0.2& 0.5& 0.2
\\
$\lambda$ & 0.2& 0.2& 0.2& 0.2& 0.2& 0.2& 0.2& 0.2
\\
\bottomrule
\end{tabular}
\caption{\label{t:9} Hyper-parameters of $\mathcal{S}_{_{\textrm{CKD}}}$.}
\end{table*}

\begin{table*}[ht!]
\centering
\begin{tabular}{l l l l l l l l l l}
\toprule
hyper-parameters & CoLA & MNLI & MRPC & QNLI & QQP & RTE & SST-2 & STS-B \\
\hline
learning rate& 5e-5	&5e-5&	5e-5&	5e-5&	5e-5&	5e-5	&5e-5&	5e-5
\\
T& 5& 20& 10& 20& 20& 10& 20& 5
\\
$\eta$& 0.2& 0.7& 0.7& 0.7& 0.7& 0.2& 0.2& 0.2
\\
$\lambda$ & 0.5& 0.2& 0.2& 0.2& 0.2& 0.2& 0.7& 0.5
\\
\bottomrule
\end{tabular}
\caption{\label{t:10} Hyper-parameters of $\mathcal{S}_{_{\textrm{ALP}}}$.}
\end{table*}

\subsection*{Hardware}
Each model is fine-tuned on a single \textit{NVIDIA 32GB V100} GPU. The fine-tuning time, based on the dataset size, can vary from a few hours to one day on a single GPU.
\end{document}